\documentclass{article}

\usepackage{PRIMEarxiv}

\usepackage[utf8]{inputenc} 
\usepackage{amsmath, amsthm, amssymb, amsfonts}
\usepackage[T1]{fontenc}    
\usepackage{hyperref}       
\usepackage{url}            
\usepackage{booktabs}       
\usepackage{amsfonts}       
\usepackage{nicefrac}       
\usepackage{microtype}      
\usepackage{lipsum}
\usepackage{fancyhdr}       
\usepackage{graphicx}       
\graphicspath{{media/}} 
\usepackage{float}

\title{Flikcer - A Chrome Extension to Resolve Online Epileptogenic Visual Content with Real-Time Luminance Frequency Analysis}

\author{
  Jaisal Kothari\\
  Student \\
  Amity International School, Saket\\
  New Delhi\\
  \And
  Ashay Srivastava \\
  Student \\
  Delhi Public School, RK Puram \\
  New Delhi
}

\begin{document}
\maketitle

\begin{abstract}
Video content with fast luminance variations, or with spatial patterns of high contrast - referred to as epileptogenic visual content - may  induce seizures on viewers with photosensitive epilepsy, and even cause discomfort in users not affected by this disease. Flikcer is a web app in the form of a website and chrome extension which aims to resolve epileptic content in videos. It provides the number of possible triggers for a seizure. It also provides the timestamps for these triggers along with a safer version of the video, free to download. The algorithm is written in Python and uses machine learning and computer vision. A key aspect of the algorithm is its computational efficiency, allowing real time implementation for public users.
\end{abstract}

\section{Introduction}

\begin{figure}[H]
  \centering
  \includegraphics[width=8cm]{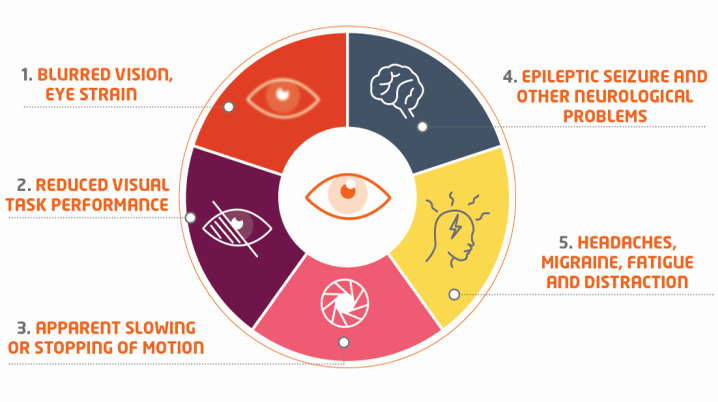}
  \caption{Harmful Effects of Flashing Video}
  \label{fig:fig1}
\end{figure}

Epilepsy is a disease that affects around 50 million people worldwide, 3-5\% of which had seizures triggered by pulsating lights, or by patterns with light and dark areas, a condition known as photosensitive epilepsy (PSE). Video content with fast luminance variations, or with spatial patterns of high contrast - referred to as epileptogenic visual content - may also induce seizures on viewers with PSE, and even cause discomfort on users not affected by this disease. To avoid this type of effects, harmful visual content should be detected before being distributed or displayed. The problem has amplified during the COVID-19 pandemic, as all education, recreational and social activities move online.

Estimates state that about 4.56 billion people have access to the internet in 2021, a number that accounts for nearly 60 percent of the world’s total population. Out of 60 percent, more than 9 out of 10 people consume video content on a daily basis. A study done by the American Academy of Children Surveys shows that children from the age of 8-12 have had a 25\% increase in average screen times from 4.5 hours a day to about 6 hours of screen time everyday.

As children and adults alike consume more and more content, they often come across unfiltered content that may include harmful things such as flickering lights. Some harmful effects of flickering lights are detailed in the diagram given above.

Our project, Flikcer, is a web app in the form of a website and chrome extension which aims to resolve epileptic content in videos. Flikcer provides a measure of how dangerously epileptic a video is, by providing the number of possible triggers for a seizure. It provides the timestamps for these triggers along with a safer version of the video, free to download.

Our algorithm has been written in python based on vector Machine Learning techniques and computer vision analyses. A key aspect of our algorithm is its computational efficiency, allowing real time implementation for the users.

\section{Research Method}
\label{sec:headings}

In order to tackle the problem, we met with Dr. Mamta, a Professor of Neurology at AIIMS (All India Institute of Medical Sciences). We also conducted online interviews with neurological scientists at Epilepsy India, and studied peer-reviewed scientific papers to understand research onto the problem we faced.
The first guidance notes related with the characterization of potential harmful video/image content were developed between 1993 and 2001 by the UK's Independent Television Commission.
The Ofcom guidelines on potential harmful flashes have been adopted by ITU-R in 2005, through recommendation ITU-R BT. A summary of these guidelines are: 

\begin{itemize}
\item A potentially harmful flash occurs when there is a pair of opposing changes in luminance of 20cd/m2 or more. This applies only when the screen luminance of the darker image is below 160cd/m2. Irrespective of luminance, a transition to or from a saturated red is also potentially harmful
\item A sequence of flashes is not permitted when both the following occur: (a) the combined area of flashes occurring concurrently occupies more than 25\% of the displayed screen area and (b) the flash frequency is higher than 3Hz
\item A sequence of flashing images lasting more than 5s might constitute a risk even when it complies with the guidelines above.
\item Rapidly changing image sequences are provocative if they result in areas of the screen that flash, in which case the same constraints apply as for flashes.
\end{itemize}

These guidelines formed the basis of our implementation.

\subsection{AI Ethics and Code}
We were inspired by Microsoft’s Responsible AI and Ethics code and have incorporated various parts of it in our product, Flicker, in the following ways:
\begin{itemize}
\item Inclusiveness - By making sure that our extension makes the Internet and its services safe for the 3\% percent of the world’s population, we also make the internet even more safe for the 97\% of the users as they can also use our extension to make their experiences on the internet far more safer.
\item Fairness - We ensure that people suffering from Photosensitive Epilepsy can enjoy the boons and joys of exploring the internet as normal users, thus allowing and granting them greater access to a wider variety of content.
\item Transparency - We give our users the right to edit the videos that they would like to establish as harmful videos, which gives users the freedom to share videos that they feel appropriate with us. This allows for greater Transparency as the user has greater control and understanding as to what video data they share with us.
\item Privacy and Security - We believe that data is the private property of the User. As our Machine learning model will require the user data to judge whether a video is safe or not, the user’s data is stored completely under their discretion and allowance.
In addition to this, the ML model stores the harmful videos on local devices and only when given allowance by the user can this list of harmful videos be uploaded to the global list of harmful videos, thus allowing greater security of user’s private data.
\item Reliability and Safety - By excessive experimentation and training of our machine learning models, we ensure that our users are given safe videos. We also made it a point to get our model evaluated by several top medical experts to ensure that our product is safe for public deployment.
\end{itemize}
Our AI system also communicates any abnormalities that it may have detected while processing content, thus allowing engineers to continuously improve our machine learning model and ensure greater standards of safety and reliability for everyone.

\section{Mathematical Modeling}
\label{sec:headings}

\subsection{Calculating Luminance}

\begin{figure}[H]
  \centering
  \includegraphics[width=4cm]{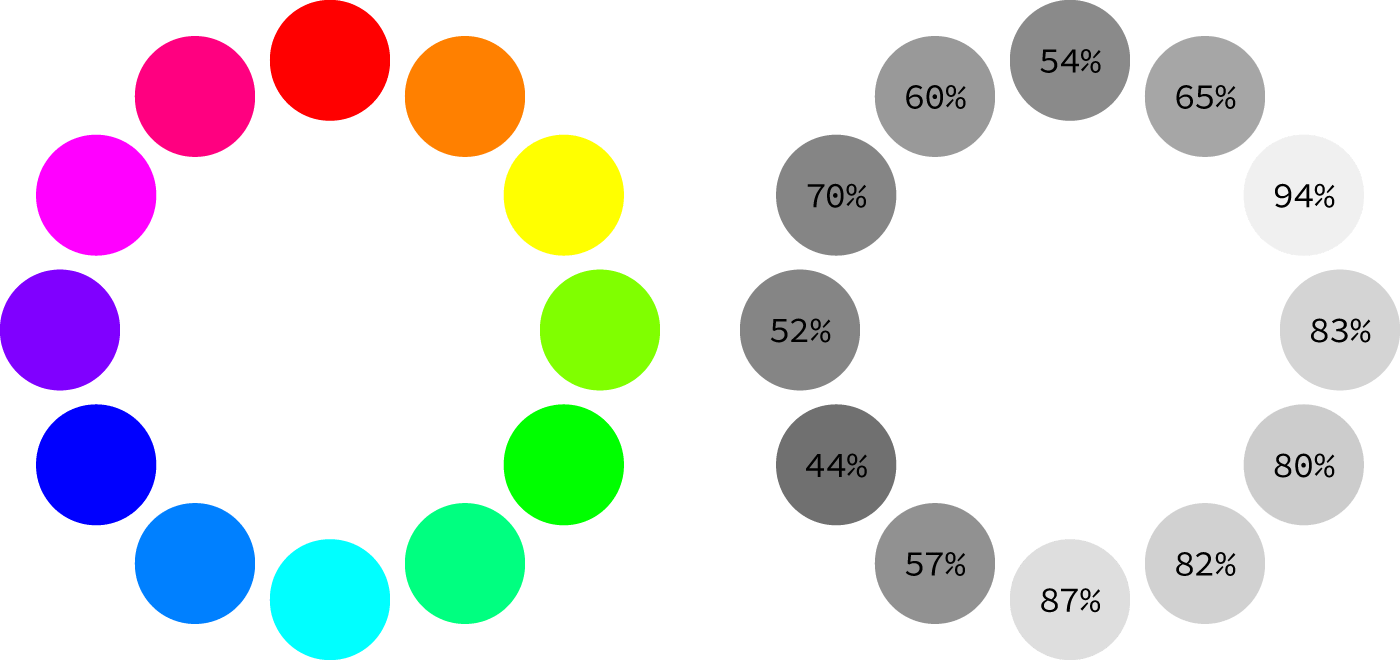}
  \caption{RGB to Luminance}
  \label{fig:fig1}
\end{figure}

Video RGB Pixel values (between 0 and 255) are converted to luminance values (cd/m2) using the following formula.

\begin{equation}
\lambda =413.435(0.002745 \Upsilon + 0.0189623)^{2.2}
\end{equation}

where $\lambda$ is the luminance value and $\Upsilon$ is the average RGB Pixel Value.

\begin{figure}[h]
  \centering
  \includegraphics[width=8cm]{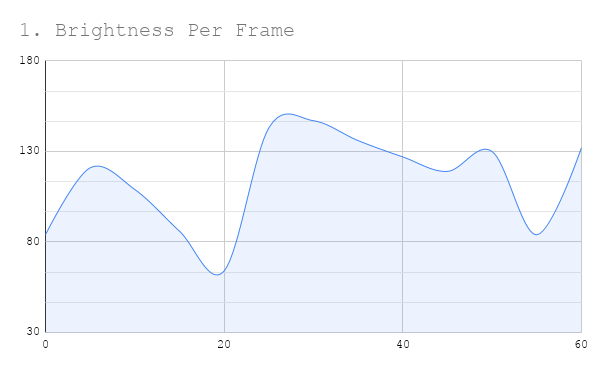}
  \caption{Brightness per Frame}
  \label{fig:fig1}
\end{figure}

\subsection{Detecting Changes in Luminance}
For each frame, compute the brightness difference (pixel by pixel) using the precedent frame

\begin{equation}
\Delta \lambda _{(x,y)}= \lambda _{n(x,y)} - \lambda _{n-1(x,y)}
\end{equation}

where $(x,y)$ are the pixel spatial coordinates and $n$ is the frame index.

\begin{figure}[h]
  \centering
  \includegraphics[width=8cm]{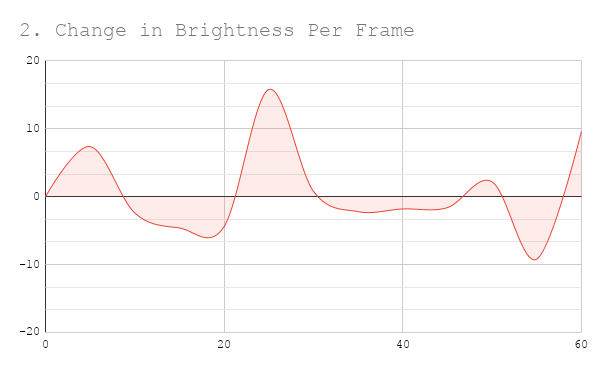}
  \caption{Change in Brightness Per Frame}
  \label{fig:fig1}
\end{figure}

\subsection{Defining a Flash}

Generate two sets, $\zeta_{n+}$ and $\zeta_{n-}$ of, respectively, the positive and negative brightness differences.

Scan $\zeta_{n+}$ and $\zeta_{n-}$, starting in each case from the highest item, and until the number of elements in the set equals the minimum number of pixels for defining a flash. This is equal to

\begin{equation}
\zeta _{s}\times K_1
\end{equation}

where $\zeta_{s}$ is the total screen area and $K_1$ is the hyperparameter for the percentage of harmful flashes on the entire screen.

\subsection{Compute Change in Brightness}

Compute the average value, $\Delta\lambda_{avg+}$ and $\Delta\lambda_{avg-}$ , of the scanned elements. $B_{n+}$ and $B_{n-}$ are the set of bins scanned in (3). If in any of the scans the number of elements is lower than $\zeta _{s}\times K_1$, the corresponding average is set to zero. The average brightness variation at frame $n$ , $\Delta\lambda_{n}$ will be given by the highest value of $\Delta\lambda_{avg+}$ and $\Delta\lambda_{avg-}$.

Compare the signs of the current and previous average brightness variation, $\Delta\lambda_{n}$ and $\Delta\lambda_{n-1}$, respectively. If the signs are equal, the brightness variation has maintained the trend, and is accumulated in an array whose size matches the frame size.

\begin{equation}
\Delta\lambda\_acc_{n(x,y)} = \Delta\lambda\_acc_{n-1(x,y)} + \Delta\lambda_{n(x,y)}
\end{equation}

The procedure is repeated from step 1 for the next video frame. If the signs are different, the brightness variation has inverted this trend, resulting in a local extreme of it.

\begin{figure}[h]
  \centering
  \includegraphics[width=8cm]{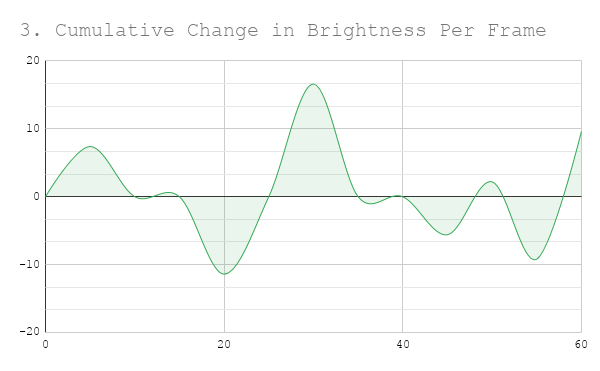}
  \caption{Cumulative Change in Brightness Per Frame}
  \label{fig:fig1}
\end{figure}

\subsection{Detecting Harmful Flash}

This local extreme is compared to a threshold value, and marked as harmful if it exceeds the allowed limit.
Once we had created our cumulative brightness difference data frame, we were ready to train the Convolutional Neural Network. The mechanism was implemented through Keras. We initially treated each individual frame of a video as independent from the others. This type of implementation caused label flickering. We, then, implemented a more advanced neural network, including LSTMs (Long Short Term Memory - this type of model is built to “remember” patterns over short and long durations) and the more general RNNs (Recurrent Neural Networks - these are used for sequential models like our data frame), helped combat this problem and led to much higher accuracy.
Adding a rolling window average based on another machine learning trained window size, enabled us to smooth out the predictions and make for a better harmful flash detection model.

\begin{figure}[h]
  \centering
  \includegraphics[width=8cm]{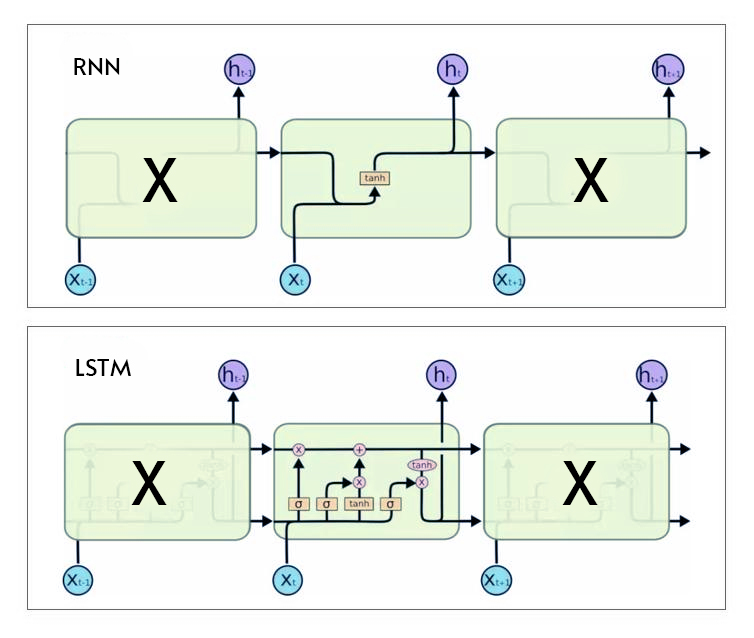}
  \caption{RNN and LSTM Models}
  \label{fig:fig1}
\end{figure}

\section{Python Model}

We programmed our algorithm in Python. Python has many helpful libraries such as numpy and opencv which can make the handling of the video and vectorized implementation of the algorithm extremely simple and effcient. We can also deploy the python algorithm on a framework like Django to connect the backend with the frontend. We use the following libraries to carry out the various tasks.

\begin{verbatim}
import cv2
import numpy as np
import os
import matplotlib.pyplot as plt
import skimage.measure
import pafy
\end{verbatim}

The next step is to obtain a video from user input in the form of a file or a YouTube URL.

\begin{verbatim}
original = [] # Stores the original video for downloading
frames = [] # Stores the new video for processing
url = str(input()) # Input YouTube URL
vPafy = pafy.new(url)
play = vPafy.getbest(preftype="mp4")
sample = play.url
vid = cv2.VideoCapture(sample)
fps = vid.get(cv2.CAP_PROP_FPS) # Get video fps
while(vid.isOpened()):
    rtrn, frame = vid.read()
    if rtrn == False:
        break
    original.append(frame)
    frame_res = cv2.resize(frame, (320,240), interpolation cv2.INTER_AREA)
    frames.append(frame_res)
vid.release()
cv2.destroyAllWindows()
\end{verbatim}

We now start the vectorized implementation of our model. This would ensure an efficient model which runs in real time. Video processing is one the most intensive tasks in computer vision and data science, therefore, computational time is of extreme importance.

\begin{verbatim}
def run(frames):
    avg_lum_frames = 413.435*(0.002745 * frames.mean(axis=3) + 0.0189623)**2.2
    change_lum = np.delete(avg_lum_frames, 0, 0) - np.delete(avg_lum_frames, -1, 0)
    pos_lum = change_lum.copy().reshape(change_lum.shape[0], -1)
    pos_lum[pos_lum < 0] = 0
    pos_lum.sort(axis=1)
    pos_lum = np.flip(pos_lum, axis=1)
    neg_lum = -change_lum.copy().reshape(change_lum.shape[0], -1)
    neg_lum[neg_lum < 0] = 0
    neg_lum.sort(axis=1)
    neg_lum = np.flip(neg_lum, axis=1)
    QUARTER = ceil(pos_lum.shape[1] / 4)
    p_avgL = pos_lum[:,:QUARTER].mean(axis=1)
    n_avgL = neg_lum[:,:QUARTER].mean(axis=1)
    table = p_avgL - n_avgL
    table[table > 0] = p_avgL[table > 0]
    table[table <= 0] = -n_avgL[table <= 0]
    return table
    
def get_fin_frame(table):
    def getSign(x):
        if x > 0 : 
            sign = "pos"
        else:
            sign = "neg"
        return sign
    fin = []
    fin_frames = []
    cum = table[0]
    fin_frame = 1

    for change in range(len(table)-1):
        if getSign(table[change]) == getSign(table[change + 1]):
            cum += table[change + 1]
            fin_frame += 1
        else:
            fin.append(cum)
        fin_frames.append(fin_frame)
        cum = table[change + 1]
        fin_frame = change + 2
    return fin_frames, fin

def get_ep_and_rm_frm(fin_frames, fin):
    ep_frm = []
    rem_frm = []
    prev = 0
    for x in range(len(fin)) :
        if abs(fin[x]) >= 20 :
            frame_inc = fin_frames[x] - prev
            prev = fin_frames[x]
    rem_frm.append(fin_frames[x])
    ep_frm.append(frame_inc)
    return ep_frm, rem_frm
    
def possible_triggers(ep_frm, fps):
    ext = 0
    score = 0
    hits = 0
    for a in range(len(ep_frm)):
        if score < fps:
            score += ep_frm[a]
            hits += 1
        else:
            if hits > 3 :
                ext += 1
            score = 0
            hits = 0
    return ext
    
def clean_video(original, frames, fps):
    table = run(frames)
    print (`\tCalculated EPFrames')
    fin_frames, fin = get_fin_frame(table)
    ep_frm, rem_frm = get_ep_and_rm_frm(fin_frames, fin)
    num_triggers = possible_triggers(ep_frm, fps)
    print("\tPossible triggers:", num_triggers)
    return num_triggers
    
def process(url):
    play, fps, original, frames = download_video(url)
    print (`Downloaded Video')
    num_triggers = clean_video(original, frames, fps)
    return num_triggers

\end{verbatim}

This code outputs the total number of triggers, along with their time stamps in the video. It also outputs an array of frame stamps, containing harmful video content.
We have now successfully detected the frames where harmful flashes occur - the possible epilepsy triggers. Now we need to remove these frames in order to create a safer version of the video.

\begin{verbatim}
from moviepy.editor import *
from itertools import groupby, count
def intervals(data, fps):
    out = []
    counter = count()
    for key, group in groupby(data, key = lambda x:x-next(counter)):
        block = list(group)
        out.append([(block[0] /fps ), (block[-1] / fps)])
    return out
frames_to_remove = [33,44,100,160]
url = ""
video = VideoFileClip(url)
fps = video.fps
level = 5
video_frames = []
extend_fps = int(fps * level)
extend_frame = 0
for i in range(len(frames_to_remove)):
    if (frames_to_remove[i] > extend_frame):
        video_frames.extend([frames_to_remove[i] + c for c in range(extend_fps)])
        extend_frame = frames_to_remove[i] + extend_fps -1
audio_frames = intervals(video_frames, fps)
audio_frames.reverse()
print(audio_frames)
print(video_frames)
for cut in range(len(audio_frames)):
    t1 = audio_frames[cut][0]
    t2 = audio_frames[cut][1]
    video = video.cutout(t1, t2)
video.write_videofile(`testetstt.mp4', codec=`libx264', audio_codec=`aac', 
temp_audiofile=`temp-audio.m4a', remove_temp=True)
\end{verbatim}

We now upload this safe video to our backend, which we will use later to create the user end product.

\begin{verbatim}
os.environ["GOOGLE_APPLICATION_CREDENTIALS"]=""
client = storage.Client()
bucket = client.get_bucket(`')
blob = bucket.blob('download/safevideo.mp4')
blob.upload_from_filename(filename=`safevid.mp4')
\end{verbatim}

\section{User-End Product}

\begin{figure}[h]
  \centering
  \includegraphics[width=\textwidth]{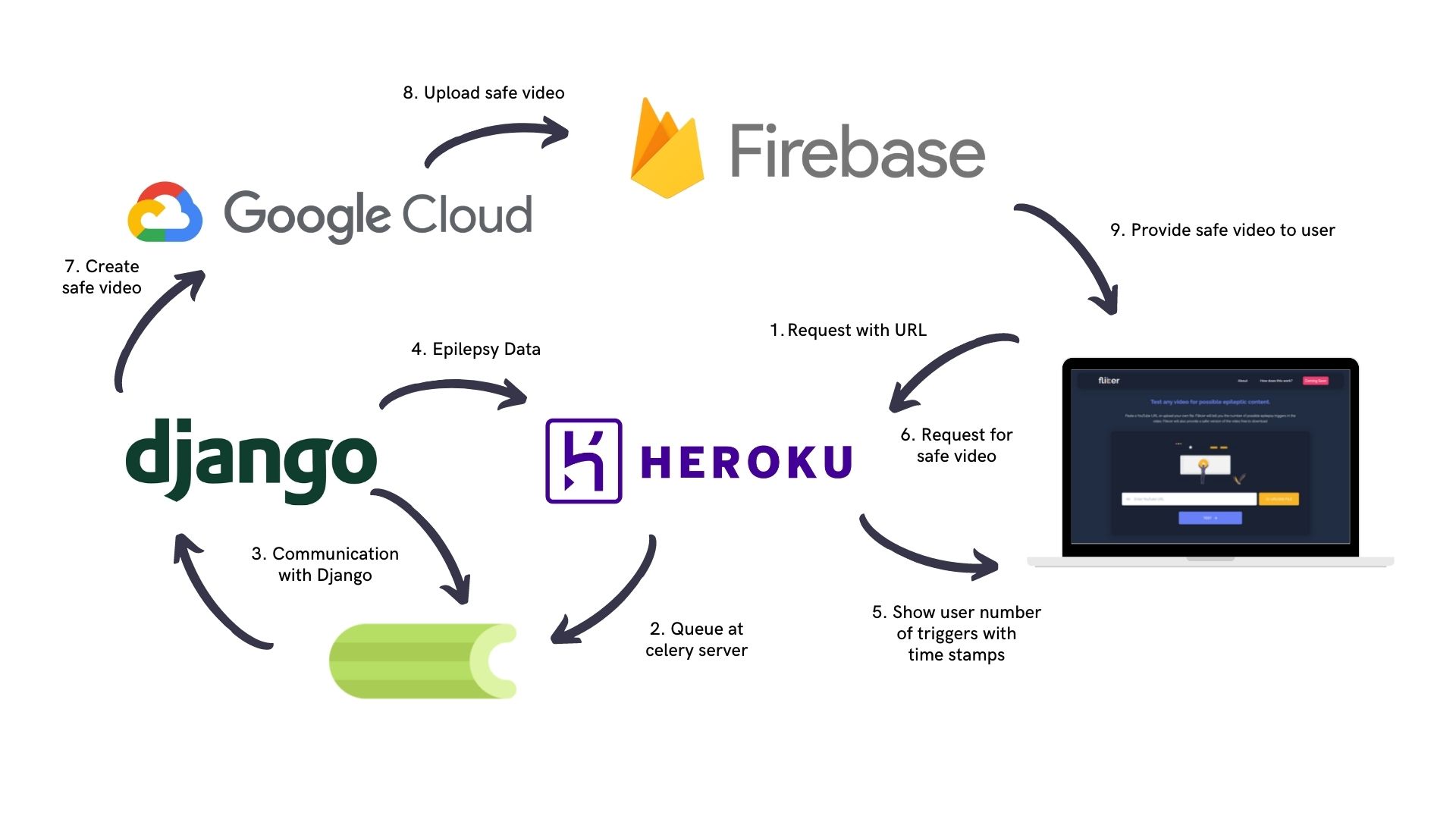}
  \caption{Product Working}
  \label{fig:fig1}
\end{figure}

In order to allow users all around the world to access the algorithm and use our python model, we decided to create a website, followed by a chrome extension. We had to keep in mind various features of a good website - User Friendly UI and UX, Optimization, Fast Loading.
We deployed our website using Heroku. All the requests are handled through a Celery server and a Django project controlling the Python.
We used Google’s Firebase to store the video files, and allow interaction with our server and the user’s device.
\\ \\
After the prototype of the website was ready, we added the various input and request functionalities. We also added information and resource tabs, since a huge part of the solution was raising awareness about the issue. Once the website was ready, we also created a chrome extension, which works directly from the browser and automatically reads the current tabs’ URLs. The website is available at \url{www.flikcerapp.com.}
\\ \\
Google accepted Flikcer as a chrome extension on its chrome web store. The chrome extension can be found at \url{https://chrome.google.com/webstore/detail/flikcer/nfjphdjibgmgeklhbbkglceokkdlpkfo?hl=en}.

\subsection{Instructions}
If you want to resolve or test any video on YouTube for possible photosensitive epilepsy triggers, simply copy and paste the url into the text field and simply, click ‘Test’.

If you want to resolve or test any video of your own for possible photosensitive epilepsy triggers, click the ‘Upload’ button and select your video file. Wait for the video to finish uploading to our servers and then, simply, click ‘Test’.

Flikcer will read your video, analyse all frames and detect the triggers. After it has done it’s analysis, Flikcer will output the number of possible triggers in that video. You can now look at the specific time stamps of those triggers by pressing the ‘View Timestamps' button.

If you want to download a safer version of your video, you can click the ‘Create Safe Video’ button. Flikcer will ask for the number of iterations you want it to run. Each iteration removes new epileptic frames that may have been created after removing the previous frames. You can input a number between 1 and 5, Flikcer will run that many iterations. Once the video has been made, you can simply click ‘Download’ to download your safe video.

\subsection{Understanding the result}

Flikcer outputs the number of possible triggers in your video. This can be seen as a measure of how dangerous a video is.
Flikcer should mainly be used to make a black and white decision - to avoid or view a video. For example, a short video with 20 triggers should be avoided. A longer video with 5 triggers can still be viewed, with a best practice of skipping 2-3 seconds around the harmful time stamps.
Flikcer does not know what kind of video you are inputting. Therefore, a flicker may actually be a flash of light or a shake of a camera. You should make your decision based on what kind of video the original input is.

\begin{figure}[H]
  \centering
  \includegraphics[width=.6\textwidth]{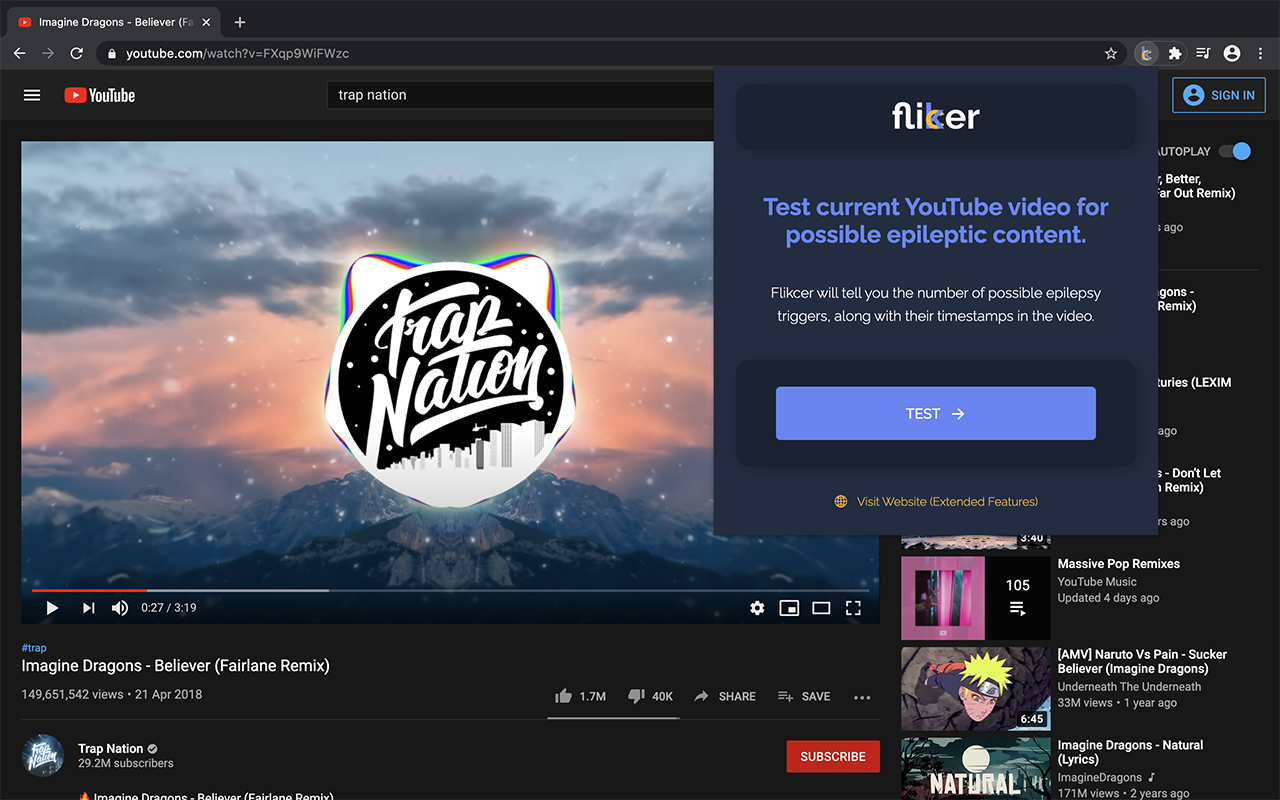}
  \caption{Chrome Extension (a)}
  \label{fig:fig1}
\end{figure}
\begin{figure}[H]
  \centering
\includegraphics[width=.6\textwidth]{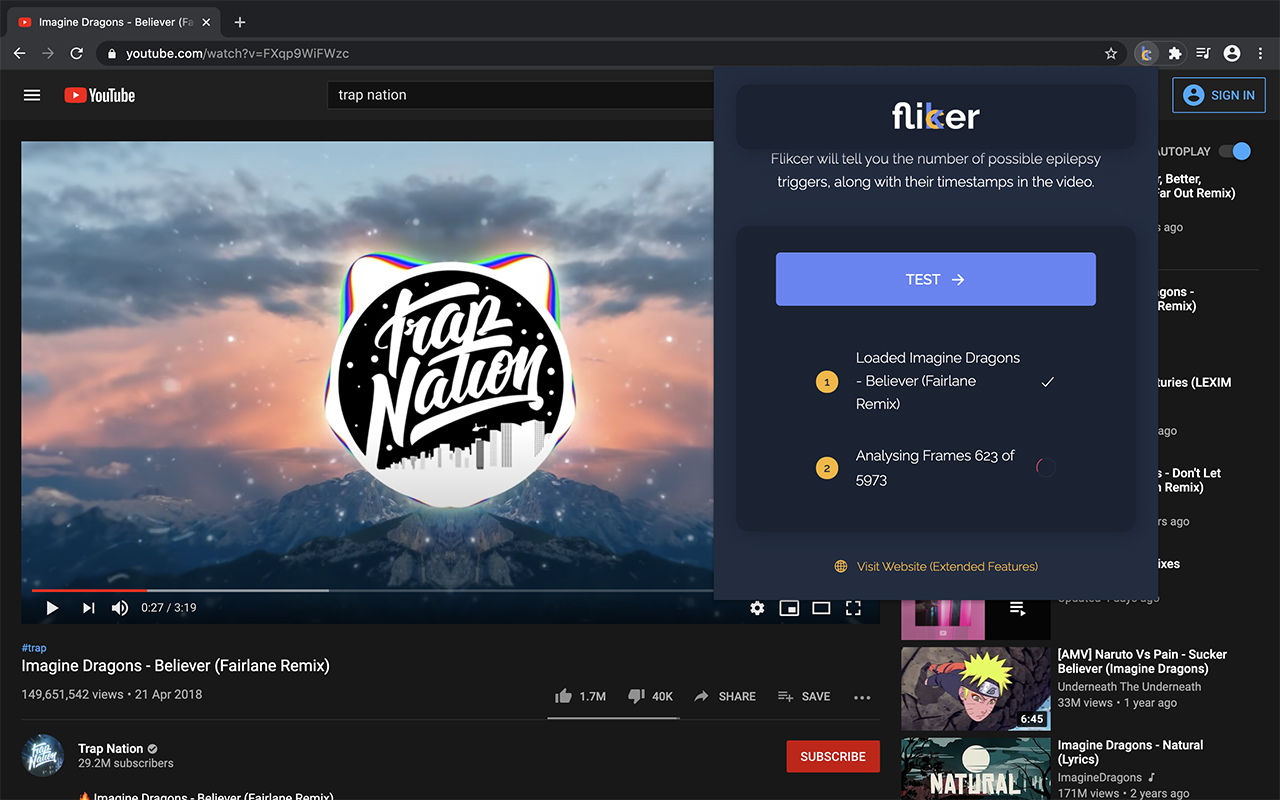}
  \caption{Chrome Extension (b)}
  \label{fig:fig1}
\end{figure}
\begin{figure}[H]
  \centering
  \includegraphics[width=.6\textwidth]{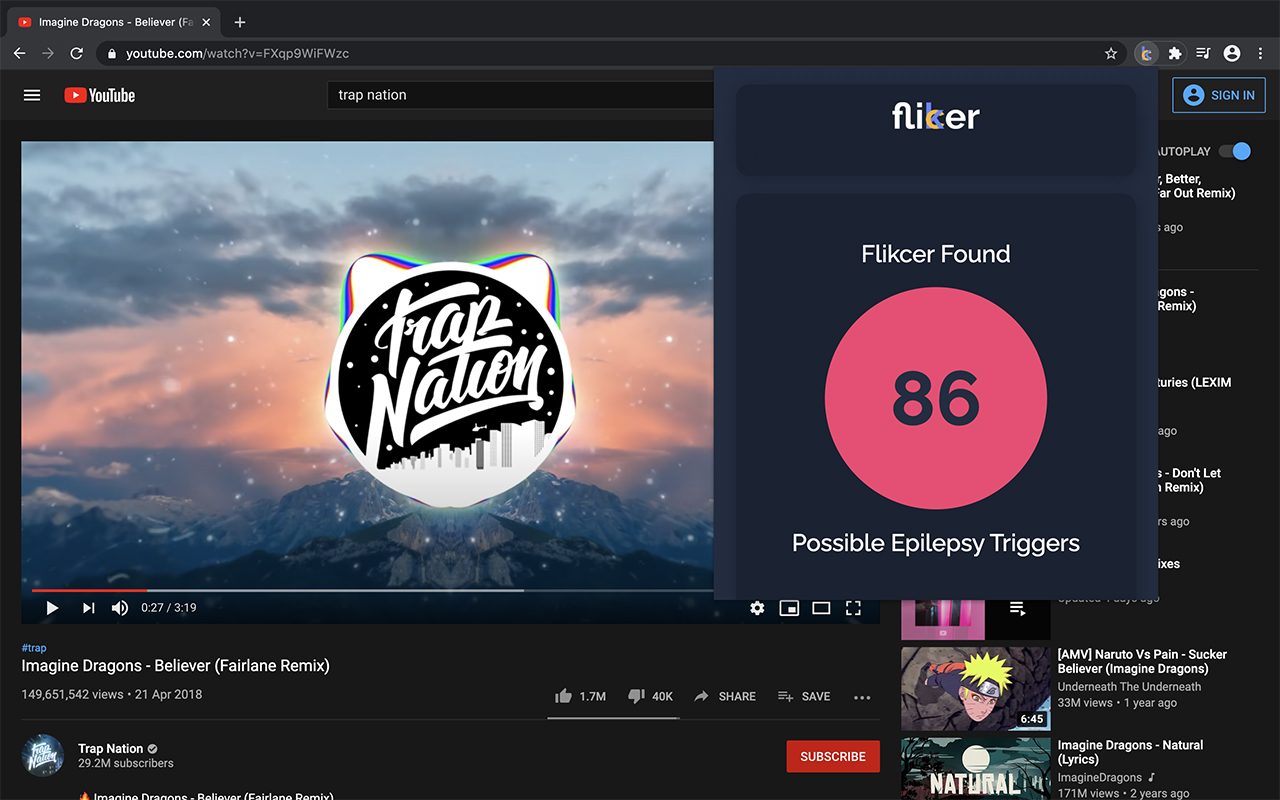}
  \caption{Chrome Extension (c)}
  \label{fig:fig1}
\end{figure}
\begin{figure}[H]
  \centering
  \includegraphics[width=.6\textwidth]{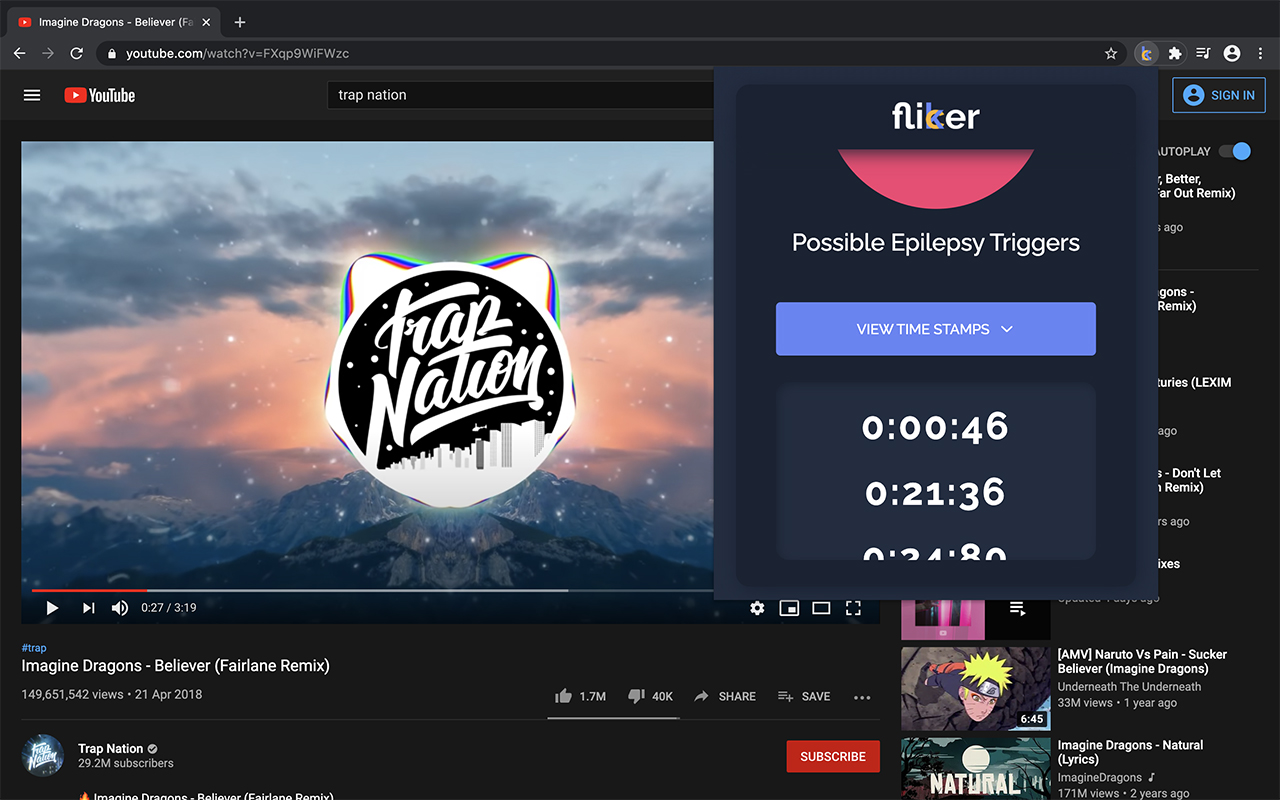}
  \caption{Chrome Extension (d)}
  \label{fig:fig1}
\end{figure}

\begin{figure}[H]
  \centering
  \includegraphics[width=.6\textwidth]{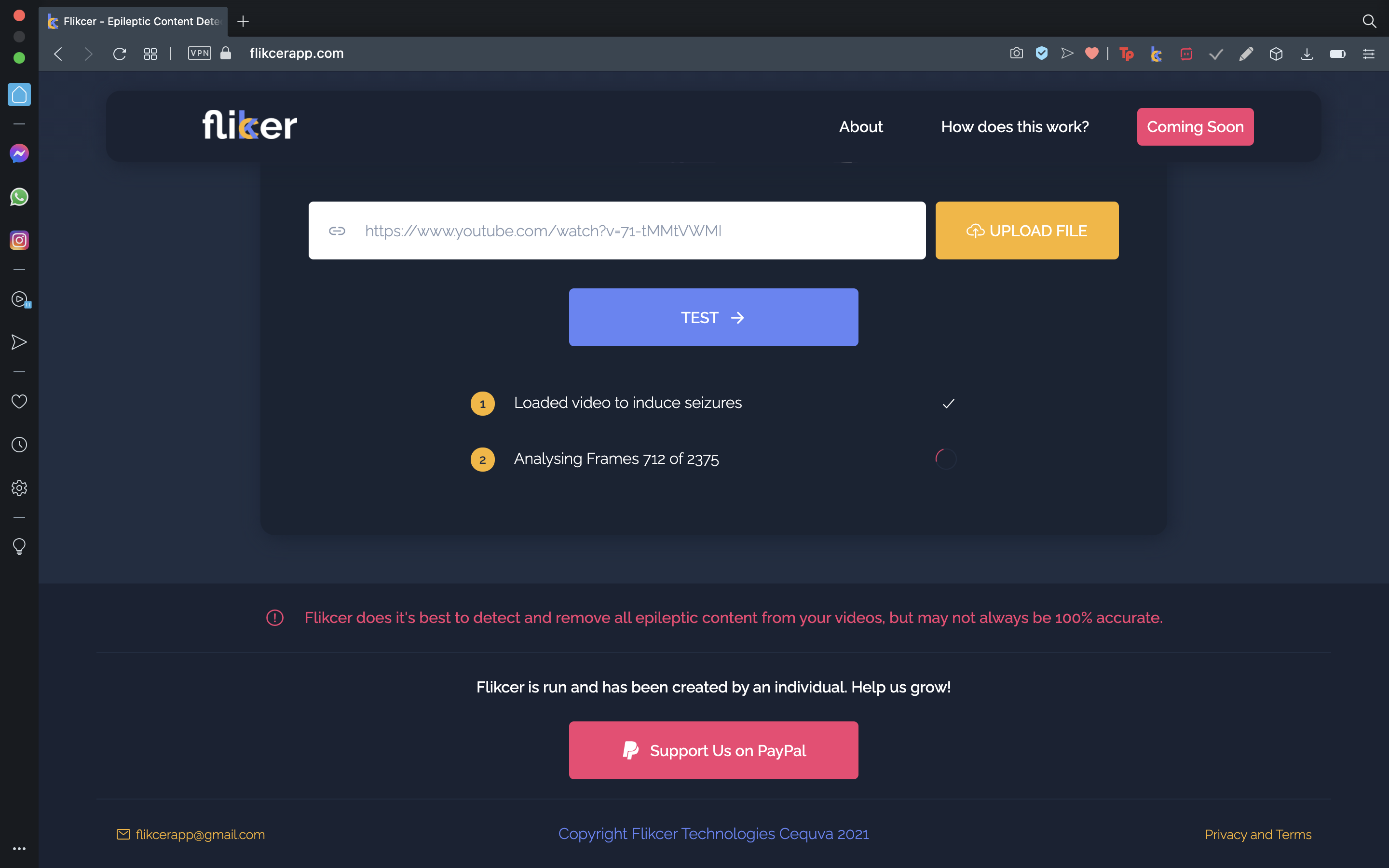}
  \caption{Website (a)}
  \label{fig:fig1}
\end{figure}
\begin{figure}[H]
  \centering
  \includegraphics[width=.6\textwidth]{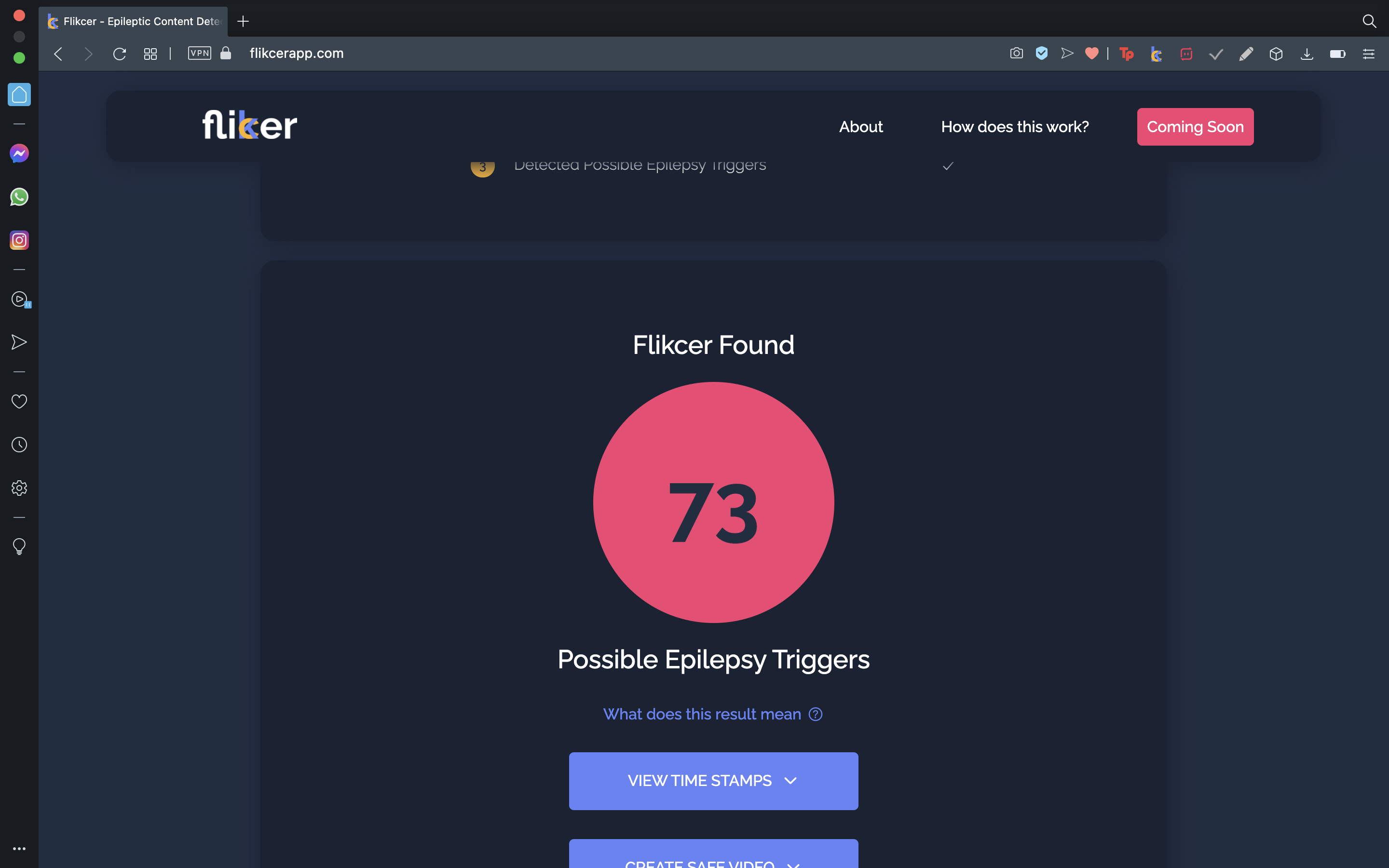}
  \caption{Website (b)}
  \label{fig:fig1}
\end{figure}
\begin{figure}[H]
  \centering
  \includegraphics[width=.6\textwidth]{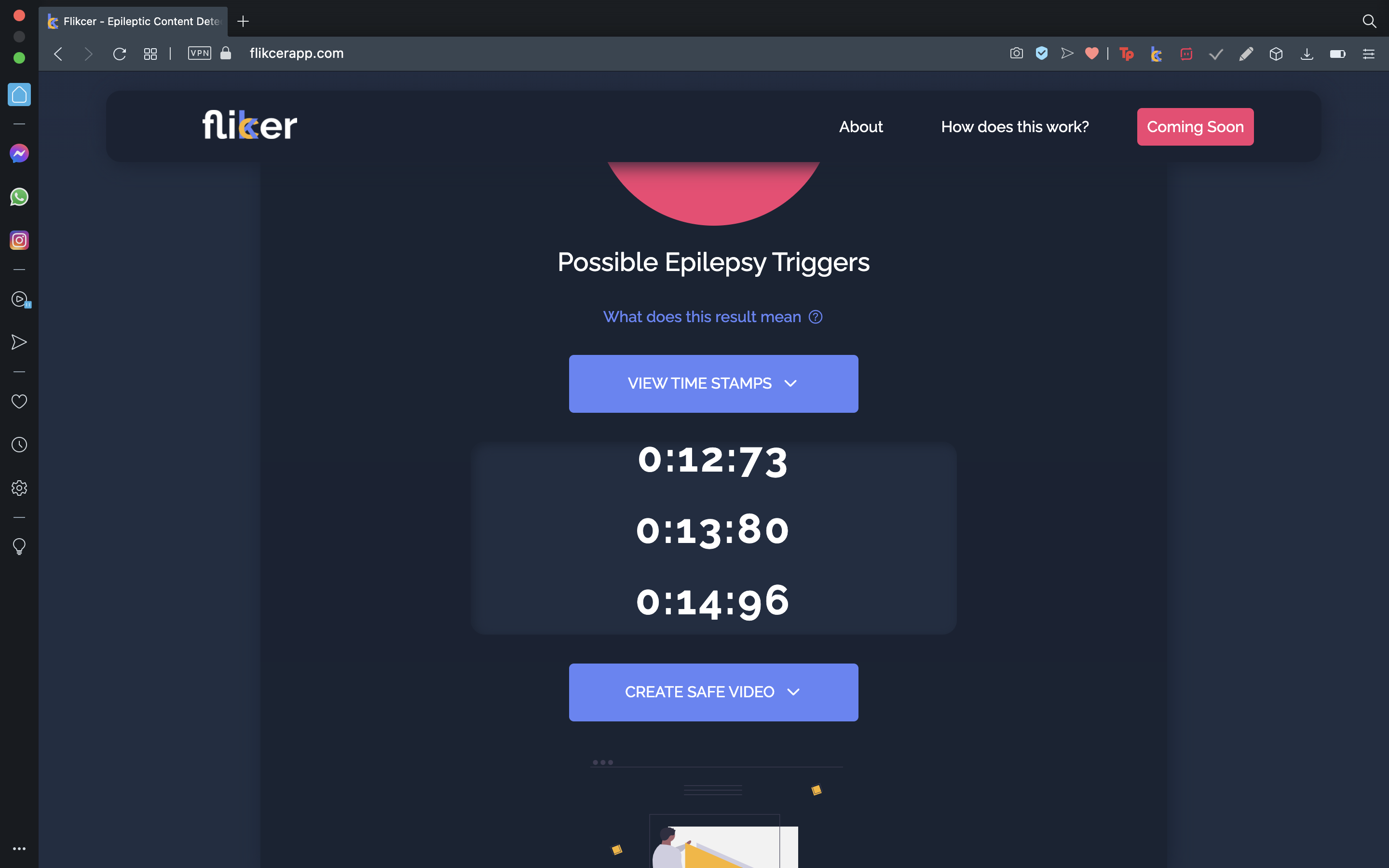}
  \caption{Website (c)}
  \label{fig:fig1}
\end{figure}
\begin{figure}[H]
  \centering
  \includegraphics[width=.6\textwidth]{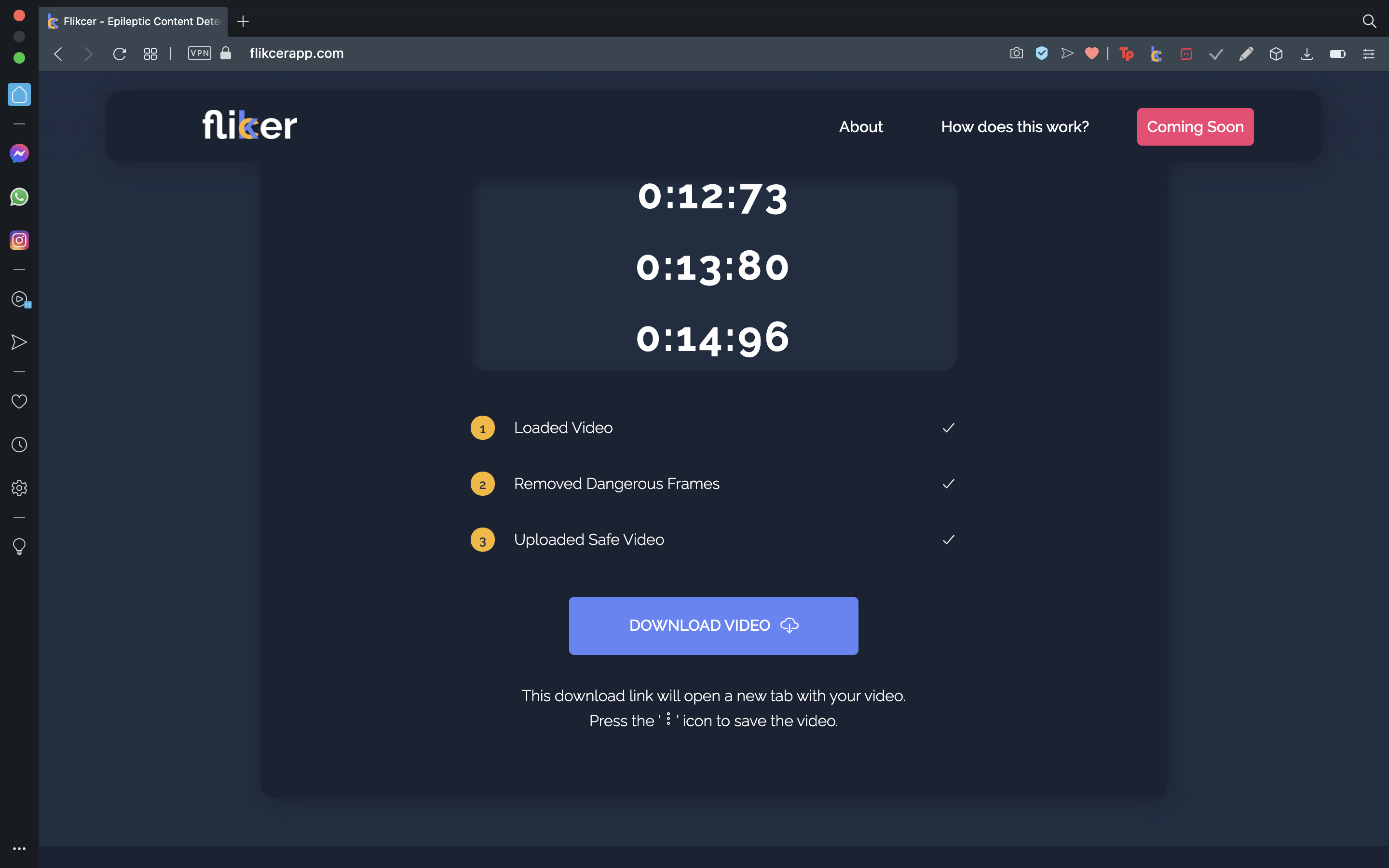}
  \caption{Website (d)}
  \label{fig:fig1}
\end{figure}

\section{Results}

\begin{table}[h]
 \caption{Testing Summary}
  \centering
  \begin{tabular}{llll}
    \toprule
    \multicolumn{4}{c}{Output}                   \\
    \cmidrule(r){3-4}
    S.No     &Video Size     &Algorithm Run Time &Detected Triggers\\
    \midrule
    1 & 2375 frames  & 27.78 s     &73\\
    2     & 544 frames & 12.36 s      &0\\
    3     & 2948 frames       & 31.98 s  &4\\
    4     & 17462 frames       & 2 min 53.02 s  &145\\
    \bottomrule
  \end{tabular}
  \label{tab:table}
\end{table}

\subsection{Overview}

As of August 2021, Flikcer has been used to resolve more than 2400 different videos by over 370 users. 7\% of these have been uploaded videos, while YouTube has been the majority source of 93\% online content. 

\subsection{Test Case 1}

Title - ‘video to induce seizures’\\
URL - \url{https://www.youtube.com/watch?v=71-tMMt VWMI}\\
This video had been made for the purpose of inducing a seizure. It contains 1 minute 19 seconds of flashing lights and rapidly changing colorful graphics. A neuroscientist would mark this video as being extremely dangerous to patients with photosensitive epilepsy as well as mildly harmful to the general population.\\
Flikcer detects 73 possible epilepsy triggers, an therefore, successfully worked in marking the video as dangerous and possibly epileptic. A patient would avoid watching this video.

\subsection{Test Case 2}

Title - ‘Nature Beautiful short video 720p HD’\\
URL - \url{https://www.youtube.com/watch?v=668nUC eBHyY}\\
This video is a short montage of nature footage. It is 20 seconds long, and contains no flashes or rapidly changing lights. It is a normal calm video. A neuroscientist would mark this video as being safe to view for patients with photosensitive epilepsy.
\\
Flikcer does not detect any possible epilepsy triggers, and therefore, successfully worked in marking the video as safe. A patient would continue watching this video, without any risk.

\subsection{Test Case 3}

Title - ‘Into The Nature - Cinematic Travel Video | Sony a6300’\\
URL - \url{https://www.youtube.com/watch?v=fEErySYq ItI}\\
This video is a 2 minute long cinematic video. Even though, overall, it seems to be a relatively safe video, with calm transitions and smooth footage, it has a few sharp cuts and flashy transitions. A neuroscientist would mark this video as being mostly safe to view for patients with photosensitive epilepsy, except for those parts where the flashes occur in the form of cinematic transitions.\\
Flikcer detects 4 possible epilepsy triggers, and therefore, successfully worked in detecting the flashes in the video. A patient would continue watching this video, but avoid watching 2-5 seconds around the various harmful timestamps outputted by the website.

\subsection{Test Case 4}

Title - ‘Avengers Endgame Final Battle with Thanos - Assemble - 4K’\\
URL - \url{https://www.youtube.com/watch?v=lHw-6A ZvZ7I}\\
This video is a 10 minute long battle scene from the movie Avengers Endgame. Even though it appears to be a normal safe to view video for the general population, fight scenes like the one in this video contain lots of effects and graphics. These range from explosions to fast cuts to sharp transitions. All of these rapidly changing flashes of light can induce seizures in patients with epilepsy. A neuroscientist would mark this video as being possibly dangerous to patients with photosensitive epilepsy, but mostly safe to the general population.\\
Flikcer detects 145 possible epilepsy triggers, and therefore, successfully worked in marking the video as harmful for epilepsy patients. With a total frame count of 17462 frames, only 0.83\% of the entire video is harmful. Therefore, an epileptic patient can easily use the website to remove those dangerous frames, and still be able to view the video, like any other person.

\section{Conclusion}

In this project, we have developed a computationally efficient, and user-friendly solution for the detection of flashing video content. This type of content is potentially harmful to people who are prone to photosensitive epilepsy and may also cause discomfort for users not affected by this disease, conditioning their quality of experience while watching TV, films, or playing video games.\\
The development of our product is also important to increase the awareness of content creators and content providers, to the need of evaluating the risk of the digital video content that people, especially teenagers, watch.\\
Flikcer allows users to input local video files as well as online videos in the form of URLs. It successfully detects and removes harmful flashes from these videos in low amounts of time, allowing for real-time use in the form of a chrome extension as well as a website. It keeps track of previously inputted videos and rapidly searches the database for previous videos, allowing for even faster results.\\
Depending on the risk factor the patient suffers from, there are options of ‘low’, ‘medium’ and ‘high’ for the frame-removal process, which is a compromise between how safe the video is and how much of the original video is left.\\
Our product also helps in creating positive awareness about photosensitive epilepsy, with resources, studies as well as general information about epilepsy and its methods of detection.

\paragraph{Recognition}
Flikcer was featured in the Times of India, India's largest newspaper, in October 2020. The story was also covered by India Today, Gadgets Now and other online newspapers. In order to check the validation of our product, we had it reviewed by Dr. Mamta Bhushan, a neuroscientist at AIIMS, Delhi. She was extremely impressed by the project and believed it to be a big step towards helping patients suffering from photosensitive epilepsy. She was willing to give us a letter of commendation as an approval of our work. In August 2021, the project won the Government of India CSIR Innovation Award - the highest award in Science and Research for high school students. We were awarded a cash prize of Rs. 50,000 to motivate further research.

\section  {References}

\begin{enumerate}
    \item A.J. Wilkins, “Visual stress”, Oxford University Press, 1995.
    \item A.J. Wilkins, J. Emmett, G.F.A. Harding, “Characterizing the patterned images that precipitate seizures and optimizing guidelines to prevent them”, Epilepsia, vol. 46, pp. 1212-1218, 2005.
    \item Arnold J Wilkins, Ann Baker, Devi Amin, Shelagh Smith, Julia Bradford, Zenobia Zaiwalla, Frank M.C Besag, Colin D Binnie, David Fish,
Treatment of photosensitive epilepsy using coloured glasses,
Seizure,
Volume 8, Issue 8,
1999,
Pages 444-449,
ISSN 1059-1311,
https://doi.org/10.1053/seiz.1999.0337.
(https://www.sciencedirect.com/science/article/pii/S1059131199903370)
\item Carreira, Lúcia \& Rodrigues, Nelson \& Roque, Bruno \& Queluz, Maria. (2015). Automatic detection of flashing video content. 2015 7th International Workshop on Quality of Multimedia Experience, QoMEX 2015. 10.1109/QoMEX.2015.7148104. 
    \item C.D. Binni, J. Emmett, P. Gardiner, “Characterising the flashing television images that precipitate seizures”, SMPTE, vol.111, pp. 32-39, 2002.
    \item G.F.A. Harding, P.M. Jeavons, “Photosensitive epilepsy”, London, MacKeith Press, 1994.
    \item G.F.A. Harding, P.F. Harding,
Photosensitive epilepsy and image safety,
Applied Ergonomics,
Volume 41, Issue 4,
2010,
Pages 504-508,
ISSN 0003-6870,
https://doi.org/10.1016/j.apergo.2008.08.005.
(https://www.sciencedirect.com/science/article/pii/S0003687008001282)
    \item Ofcom, "Guidances Notes, section 2: harm and offence", Issue 10, 23
July 2012.
    \item R. Fisher, G.F.A. Harding, G. Erba, G.L. Barkley, A.J. Wilkins, “Photic- and pattern-induced seizures: a review for the Epilepsy Foundation of America Working Group”, Epilepsia, vol. 46(9), pp. 1426-1441, 2005.
    \item S. Clippingdale, H. Isono, " Photosensitivity, broadcast guidelines and video monitoring", IEEE International Conference on Systems, Man, and Cybernetics, Vol. 2, pp. 22 - 27, 1999.
\end{enumerate}

\end{document}